\documentclass[10pt, a4paper]{article}
\usepackage{lrec}
\usepackage{multibib}
\newcites{languageresource}{Language Resources}
\usepackage{graphicx}
\usepackage{tabularx}
\usepackage{soul}
\usepackage{multirow}
\usepackage{textcomp}
\usepackage{CJKutf8}
\usepackage{todonotes}

\usepackage{mathtools}
\usepackage{epstopdf}
\usepackage[T2A,T1]{fontenc}
\usepackage[utf8]{inputenc}
\usepackage[bulgarian,english]{babel}
\usepackage{hyperref}
\usepackage{xstring}
\usepackage{booktabs}
\usepackage{xcolor}
\usepackage{xspace}

\usepackage{tempora}

\usepackage{pbox}

\usepackage{subcaption}
\usepackage{multirow}
\usepackage{array}
\newcolumntype{L}[1]{>{\raggedright\let\newline\\\arraybackslash\hspace{0pt}}p{#1}}
\newcolumntype{C}[1]{>{\centering\let\newline\\\arraybackslash\hspace{0pt}}p{#1}}
\newcolumntype{R}[1]{>{\raggedleft\let\newline\\\arraybackslash\hspace{0pt}}p{#1}}

\newcommand{\secref}[1]{\StrSubstitute{\getrefnumber{#1}}{.}{ }}
\newcommand{\zh}[1]{\begin{CJK}{UTF8}{gbsn}#1\end{CJK}}
\newcommand{\bg}[1]{\selectlanguage{bulgarian}#1\selectlanguage{english}}
\renewcommand{\thesection}{\arabic{section}}
\renewcommand{\thesubsection}{\arabic{section}.\arabic{subsection}\xspace}

\newcommand{\bleu}{\textsc{Bleu}\xspace}

\title{Document Sub-structure in Neural Machine Translation}

\name{Radina Dobreva$^1$, Jie Zhou$^2$, Rachel Bawden$^1$}

\address{$^1$School of Informatics, University of Edinburgh, 
         10 Crichton Street, Edinburgh, United Kingdom\\
         $^2$Alibaba Group, 969 West Wen Yi Road, Hangzhou, China
         \\
         rdobreva@ed.ac.uk, zj236040@alibaba-inc.com, rbawden@ed.ac.uk}

\abstract{
Current approaches to machine translation (MT) either translate sentences in isolation, disregarding the context they appear in, or model context at the level of the full document, without a notion of any internal structure the document may have. In this work we consider the fact that documents are rarely homogeneous blocks of text, but rather consist of parts covering different topics. Some documents, such as biographies and encyclopedia entries, have highly predictable, regular structures in which sections are characterised by different topics. We draw inspiration from \newcite{lw2014} who use this information to improve statistical MT and transfer their proposal into the framework of neural MT. We compare two different methods of including information about the topic of the section within which each sentence is found: one using side constraints and the other using a cache-based model. We create and release the data on which we run our experiments -- parallel corpora for three language pairs (Chinese-English, French-English, Bulgarian-English) from Wikipedia biographies, which we extract automatically, preserving the boundaries of sections within the articles.    \\ \newline \Keywords{machine translation, document structure, corpus creation, context, Wikipedia, parallel corpus} }

\begin{document}

\maketitleabstract

\section{Introduction}

% Introduce general topic of sub-structure modelling
Considerable progress has been made in machine translation (MT) thanks to the use of neural MT (NMT). While most NMT systems translate at the level of individual sentences, following similar practices in statistical MT (SMT), there has been significant interest in recent years in using document context to improve translation \cite{hardmeier_discourse_2014,bawden_thesis_2018,wang_thesis_2019}. %Techniques include domain adaptation \cite{} and document-level translation \cite{}.
However the intermediate level of the internal structure of documents, particularly for documents with regular sub-structure, could also provide useful information to improve MT.%, and this is what we explore here for NMT.

\begin{figure}
    \centering
    \includegraphics[width=1\linewidth]{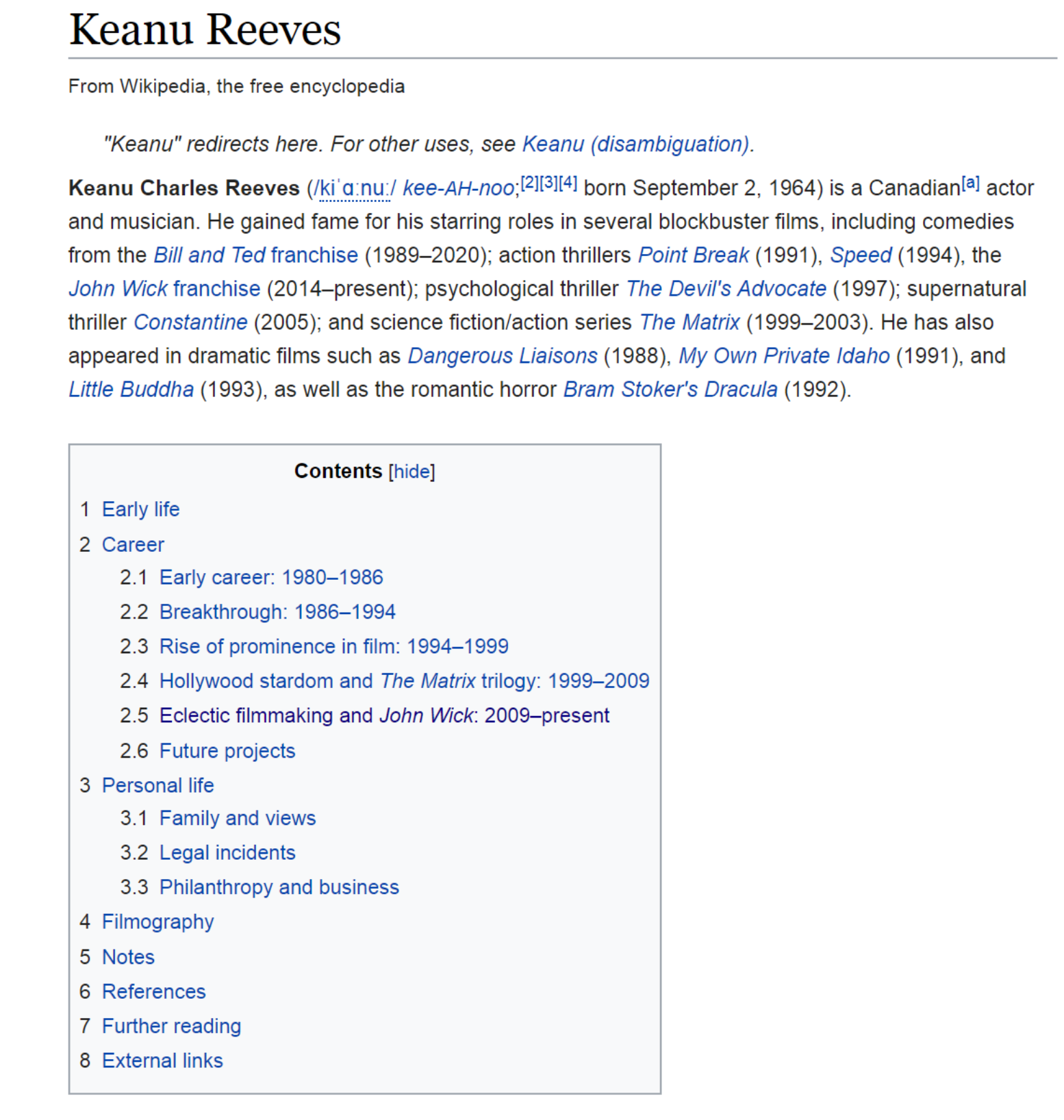}
    \caption{An extract of a Wikipedia biography article, in which the document structure is summarised.}
    \label{fig:biography-example}
\end{figure}

% What is special about these kinds of documents?
Documents are rarely without internal structure, and certain document types (e.g.~biographies, scientific articles and encyclopedia entries) are characterised by a more well-defined and regular structure than others. In such documents, the structure is explicitly defined by sections associated with headings dealing with different aspects of the main subject. These sections, many of which can be found across multiple documents, are often associated with specific vocabularies or even grammatical patterns. Figure~\ref{fig:biography-example} illustrates the classic structure of a Wikipedia article, with section headings (e.g.~``early life'', ``career'', ``personal life'', etc.) that are likely to be found in other articles concerning high-profile people. This regularity in document sub-structure could be beneficial to the MT of such documents by providing additional information about the type of vocabulary used in different sections.

% What do we do in this article? And what is different from previous work?
Exploiting document sub-structure in MT has previously been studied by 
\newcite{lw2014} for SMT. They use topic models to integrate section information into a cache-based system and see improvements when applying their method to the translation of Wikipedia biographies for French$\rightarrow$English. However MT techniques have since changed considerably, and the quality of MT has improved with the introduction of NMT. The change in paradigm provides us with new methods of integrating external information.

In this article, we draw inspiration from \cite{lw2014} to explore the effect of using predictable document structure to improve NMT. 
As in their work, we focus on the domain of Wikipedia biographies, which are segmented into sections covering different aspects of the life of the person they are describing -- their childhood, career, personal life, old age, etc.
We automatically create datasets of parallel Wikipedia biographies in three language pairs - French-English (Fr-En), Bulgarian-English (Bg-En) and Chinese-English (Zh-En) - preserving document and document sub-structure information. We also collect monolingual Wikipedia biography corpora in the four languages.\footnote{Available at \url{https://github.com/radidd/Doc-substructure-NMT} under a CC-BY-SA 3.0 licence.}

%In addition, 

To test the usefulness of exploiting document substructure for NMT, we conduct experiments to compare methods using topic information at the level of document sections. As in \cite{lw2014}, we use topic modelling to model article sections. %as a distribution over the vocabulary. 
We compare two methods of integrating this information into NMT. The first uses side constraints \cite{sennrich2016} and involves prepending topic information to source sentences. The second, adapted from \cite{kuang2018}, uses caches containing relevant vocabulary. It is similar to the approach used by \newcite{lw2014} but applied to NMT. We test these methods on three language directions (Fr$\rightarrow$En, Zh$\rightarrow$En, Bg$\rightarrow$En).

% Summary of the contributions in list form
Our main contributions can be summarised as follows:

\begin{itemize}
    \item The automatic creation of parallel and monolingual datasets of Wikipedia biographies with document sub-structure information for 
    Fr-En, Bg-En and Zh-En
    \item Experiments comparing two methods of exploiting document sub-structure in NMT, applied to three language directions: Fr$\rightarrow$En, Zh$\rightarrow$En and Bg$\rightarrow$En.
\end{itemize}

The remainder of the paper is organised as follows. Section~\ref{sec:background} presents relevant work on modelling document structure and background on the approaches we compare. Section~\ref{sec:datasets} describes the collection and processing of our datasets. In Section~\ref{sec:integrating-structure} we present two methods of integrating document structure information. Section~\ref{sec:experiments} gives the experimental setup and results, and Section~\ref{sec:analysis} presents analysis of these results. Finally, in Section~\ref{sec:conclusion} we provide a conclusion and ideas for further research.

\section{Related Work}\label{sec:background}
    
The exploitation of document sub-structure has previously been studied in the context of SMT by \newcite{lw2014}. The authors focus on the domain of Wikipedia biographies, presenting domain-adapted MT models. As they discuss, the biography domain is interesting because it consists of documents with a regular structure. Most biographies consist of sections that discuss topics such as their early life, career (which can fall into many categories based on the person), personal life and later life. To demonstrate the usefulness of document structure for MT, they compare models using document-wide context with models using context at the level of individual sections.
% Use topic models to get information
%
They use two caches to pass information about a sentence's topic and its preceding context to the MT model. In their structured model they load the topic cache with words that are specific to the given section, as opposed to the whole document. They also clear and reload the caches at section boundaries instead of at document boundaries. They show that the structured model has an advantage over the model that treats documents as a whole, with topic information found to be particularly useful.
    
While current work in NMT has not made use of document structure, there have been many efforts to supply NMT models with document-level context to improve translation \cite{voita2018,maruf2019,zhang2018,miculicich2018}. One way to integrate structural information into NMT is to adapt approaches to document-level NMT to consider section boundaries.

Of particular relevance is the work of \newcite{kuang2018}, who supply contextual information using caches. Their model involves two caches, a topic cache and a dynamic cache, which contain words that are important for the document-level context. The topic cache consists of words related to the document's topic, while the dynamic cache is updated to contain the translations of previous sentences within the same document and the current sentence up to the current time step. Using both the topic and the dynamic cache provides the model with document-wide information ensuring consistency, as well as information from the preceding context which allows for better coherence. %An advantage of this type of model is that the information supplied to the model can be controlled. For example, the topic cache contains words relevant to the topic of the text even though these words may not be in the preceding context. %The dynamic cache can be updated with specific words from the previous context - \newcite{kuang2018} only load content words as opposed to function words such as prepositions and conjunctions.
    
The topic information used in \cite{lw2014} and \cite{kuang2018} is supplied using topic models that identify the most important words for any given topic. The method they use is Latent Dirichlet Allocation (LDA) \cite{blei2003}. It is important to note that topic modelling can be done on entire documents as in \cite{kuang2018} or on sections within documents (as in this paper). In both cases, the text is treated as a ``bag of words'' and the important information comes from the frequency of the words in the segments. That is, topic modelling done on sections within documents will learn topics that are relevant to the text in the sections. Furthermore, while topic models treat documents as a mixture of topics with different probabilities, for downstream tasks it can be useful to only consider one or several of the most probable topics. For instance, \newcite{lw2014} only consider the topic with the highest probability for each section of a document.

\section{Datasets}\label{sec:datasets}
    
% training and dev data sizes
\begin{table*}[!ht]
\centering\small
\scalebox{0.95}{
\begin{tabular}{lrrrrrr}
\toprule
\multirow{2}{*}{}  & \multicolumn{2}{c}{Zh-En}                          & \multicolumn{2}{c}{Fr-En} & \multicolumn{2}{c}{Bg-En} \\ 
& training & validation & training & validation & training & validation \\
\midrule
Total \#documents & 693 & 75 & 23,648 & 274 & 1,899 & 149 \\
Total \#sentences & 26,505 & 2,020 & 179,270 & 2,002 & 29,348 & 2,004 \\
Average \#sentences/section (source) & 7.54 & 6.14 & 3.33 & 3.08 & 5.35 & 5.09 \\
Average \#sentences/section (target) & 7.22 & 5.87 & 3.29 & 3.10 & 4.14 & 4.00 \\
Average \#sections/document (source) & 5.07 & 4.38 & 2.28 & 2.37 & 2.89 & 2.64                           \\
Average \#sections/document (target) & 5.29 & 4.59 & 2.30 & 2.35 & 3.74 & 3.36 \\
\bottomrule
\end{tabular}}
\caption{Parallel training and validation data statistics. Source refers to one of \{Zh, Bg, Fr\} and target refers to En.} 
\label{tab:bilingual_data_trv}
\end{table*}

% test data sizes
\begin{table}[!ht]
\centering\small
\scalebox{0.96}{
\begin{tabular}{lrrrrrr}
\toprule
Lang. pair & \multicolumn{2}{c}{Zh-En} & \multicolumn{2}{c}{Fr-En} & \multicolumn{2}{c}{Bg-En} \\ 
Original lang. & En & Zh & En & Fr & En & Bg \\
\midrule
Total \#docs  & 22 & 30 & 78 & 82 & 154 & 25 \\
Total \#sents  & 1147 & 875 & 1130 & 1198 & 2539  & 273 \\
Ave. \#sents/sec & 9.89 & 6.58 & 4.56 & 4.83 & 5.37 & 3.41 \\                          
Ave. \#secs/doc & 5.27 & 4.43 & 3.18 & 3.02 & 3.06 & 3.2 \\
\bottomrule                     
\end{tabular}}
\caption{Parallel test set statistics. We split the set in two, depending on which language was the original text's language (as opposed to the translation).}
\label{tab:bilingual_data_test}
\end{table}
    
% monolingual data sizes
\begin{table}[!ht]
\centering\small
\scalebox{0.9}{
\begin{tabular}{lrrrr}
\toprule
& Zh & Fr & Bg & En \\ 
\midrule
Ave. \#sents/sec & 7.05 & 5.55 & 5.99 & 7.39 \\
Ave. \#secs/doc & 3.29 & 2.82 & 3.06 & 3.97 \\
Total \#docs & 68,433 & 167,484 & 56,275 & 99,106  \\ 
Total \#sents & 1,586,194 & 2,619,842 & 1,029,626 & 2,904,641 \\
\bottomrule
\end{tabular}}
\caption{Monolingual corpus statistics.}
\label{tab:monolingual_data}
\end{table}
    
We follow \newcite{lw2014} in using Wikipedia biographies to test whether section information can be useful for NMT. We agree that data from Wikipedia is particularly illustrative of regularly structured documents as it contains separate sections with section headings. It is reasonable to assume that some sections within different biographies might share similar vocabulary, while other sections in the same documents may be very different. For instance, a biography of a composer and one of a basketball player may both contain information about their early years, but the rest of the documents may be different. Therefore, providing more information about the recurring topic may be useful, as opposed to providing only document-level information which may be too general. 

We use a similar method to \newcite{lw2014} for data creation. However, we extend the process to cover three language pairs (Fr-En, Zh-En, Bg-En), and considerably more data,\footnote{\newcite{lw2014} use a dataset of 1000 monolingual articles for each language and parallel development and test sets of 15 and 30 articles respectively.} which is necessary as NMT models require more data to be trained effectively.

Data was collected and processed using the following steps, which are described in more detail below:
%Both monolingual and parallel data was extracted as follows:
\begin{enumerate}
\item Extraction of biographies from Wikipedia dumps and separation into parallel and monolingual data.
\item Text extraction, sentence-splitting but with preservation of document and section structure.
\item Sentence-level alignment of parallel data.
\item Cleaning of aligned parallel articles to avoid very long sentences or mismatches between source and target length using Moses scripts \cite{koehn-etal-2007-moses}.
\item Division of data into training, validation and test sets %(cf.~Table~\ref{tab:bilingual_data_trv} for statistics on these sets).
\end{enumerate}

\paragraph{Extraction of biographies}
Relevant Wikipedia articles were obtained by filtering Wikimedia dumps using their metadata. We extracted only biographies using the category of the articles with keywords related to people (e.g.~``person'', ``writer'', ``politician'') and using the presence of a ``Biography'' section as indicators.
Parallel biography data was obtained by selecting articles that were indicated in the metadata as translations (where either language was the original).
Monolingual data contains all biography data for a given language, but does not include the parallel test sets.

\paragraph{Text extraction and sentence splitting} We extracted the text from the articles, splitting it into sentences and preserving information about the document and section each sentence belongs to, as well as sentence order in the text.

\paragraph{Alignment of parallel data}
Sentence-level alignment is not trivial, especially since Wikipedia is open for anyone to edit and articles can change significantly after being translated. This is illustrated in Figure~\ref{fig:ref-examples}, which shows some examples of non-exact sentence alignment due to non-exact translation or post-editions. Some degree of such noise is therefore expected in the data. We attempt to reduce it as much as possible by filtering based on automatic alignment scores.
For Fr-En, we used \textsc{Hunalign} \cite{hunalign}, while for Zh-En and Bg-En we used previously trained MT models to translate the source text into English and align the sentences based on similarity for Zh-En and BLEU score for Bg-En using \textsc{Bleualign} \cite{sennrich-volk-2011-iterative}. Future versions of the corpus can explore how to further reduce the level of noise left in the data.
%\textcolor{red}{Mention something about alignment of the sections - or the mismatch in the sections if there is one! An example could be interesting here.}

\begin{figure*}[!ht]
    \centering\small
    \scalebox{0.93}{
    \begin{tabular}{p{5.3cm}p{5.4cm}p{6.3cm}}
    \toprule
     Source & Source gloss & Reference \\
    \midrule
    %\multirow{2}{*}{Fr} 
    Ils se marient le 30 juin 1894 à Paris. & `They married on 30 June 1894 in Paris.' & They married on 30 June 1894 in Paris \textbf{and had two daughters.} \\
    Le 11 décembre 2013, \textbf{elle} est nommée déléguée générale du Québec à New York. & `On 11 December 2013, \textbf{she} was appointed Quebec Delegate General in New York' &	On 11 December 2013 \textbf{Poirier} was appointed Quebec Delegate General in New York.\\ 
    % \textbf{C'est ce qui lui a valu} le surnom de Mathiara, talibé Almoudo. & `\textbf{This is what earned him} the nickname Mathiara talibé Almoudo.' &	\textbf{It was as a result of these studies that he received} the nickname Mathiara talibe Almoudo.\\
    \midrule
    %\multirow{2}{*}{Zh} & 
    \zh{\textbf{毛贻昌}选中了罗一秀。} & `\textbf{Mao Yichang} selected Luo Yixiu.' &  \textbf{He} selected Luo Yixiu \textbf{in either late 1907 or 1908}. \\
    \zh{1908年，\textbf{罗一秀与毛泽东}举办了婚礼} & `In 1908, \textbf{Luo Yixiu and Mao Zedong} held a wedding.' & The wedding took place in 1908. \\ 
    % \zh{林布兰的顶峰之作当属肖像画包括自画像以及取自圣经内容的绘画} & `Rembrandt's forte is portraiture, including self-portraits and paintings from the Bible' &  Rembrandt's portraits of his contemporaries, self-portraits and illustrations of scenes from the Bible are regarded as his greatest creative triumphs. \\
    \midrule
    %\multirow{2}{*}{Bg} 
    \vspace{-0.32cm}\bg{Първият му гол в Серия А е на 16 септември 2012 г. срещу Парма.} & `His first goal in Serie A is on 16 September 2012 against Parma.'	& On 16 September 2012, he scored his first goal in Serie A, \textbf{after entering as a substitute for Edinson Cavani in a 3-1 home win over Parma}. \\
    % \bg{Превратът е посрещнат враждебно от всички политически партии, представители на които го осъждат в различна степен.} & `The coup was received with hostility by all political parties, whose representatives condemned it to different degrees.' & \textbf{He abolished all political parties and trade unions}. \\ 
    \vspace{-0.33cm}\bg{Два пъти, \textbf{през 1962 и 1967 година}, получава званието ``Герой на социалистическия труд''.} & `Twice, \textbf{in 1962 and 1967}, he received the title ``Hero of socialist labour''.' & \textbf{Georgiev} was twice awarded the title Hero of Socialist Labour. \\
    \bottomrule
    \end{tabular}}
    \caption{Examples of non-exact or paraphrased references in the biography data, most likely due to translation divergences and/or subsequent modification. Differences between source and target sentences are indicated in bold.}
    \label{fig:ref-examples}
\end{figure*}

% 毛贻昌选中了罗一秀。	He selected Luo Yixiu in either late 1907 or 1908.
% 1908年，罗一秀与毛泽东举办了婚礼。	The wedding took place in 1908.
% 林布兰的顶峰之作当属肖像画包括自画像以及取自圣经内容的绘画。	Rembrandt's portraits of his contemporaries, self-portraits and illustrations of scenes from the Bible are regarded as his greatest creative triumphs.

% Първият му гол в Серия А е на 16 септември 2012 г. срещу Парма.	On 16 September 2012, he scored his first goal in Serie A, after entering as a substitute for Edinson Cavani in a 3-1 home win over Parma.
% Превратът е посрещнат враждебно от всички политически партии, представители на които го осъждат в различна степен.	He abolished all political parties and trade unions.
% Два пъти, през 1962 и 1967 година, получава званието "Герой на социалистическия труд".	Georgiev was twice awarded the title Hero of Socialist Labour.

\paragraph{Dataset partitions}
Parallel data is divided into training, validation and test sets.
For each language pair, we created two test sets, based on the original language the articles were written in (e.g.~for Bg-En, one test set contains articles originally written in English and translated into Bulgarian, and the other contains articles originally written in Bulgarian and translated into English).\footnote{It is interesting to separate out these two translation directions, as the translation direction can have an effect on ease of translation due to the ``translationese'' effect \cite{zhang-toral-2019-effect,graham_translationese_2019}} Statistics for the training and validation sets are given in Table~\ref{tab:bilingual_data_trv} and for the test sets in Table~\ref{tab:bilingual_data_test}. Monolingual data statistics are given in Table~\ref{tab:monolingual_data}. Parallel training data sizes range from 29,348 sentences for Bg-En to 179,270 sentences for Fr-En. More monolingual data is available, from 1,029,626 sentences for Bulgarian to 2,904,641 sentences for English.

\section{Integrating Document Sub-structure}\label{sec:integrating-structure}

We compare two methods of integrating document structure information, both relying on the assumption that different sections of the text cover different topics. Here, we consider different sections to be those that are delimited by headings (whether these are section, subsection or subsubsection headings), which creates a flat hierarchy of delimited areas of text.
In both cases, we follow \newcite{lw2014} in using topic modelling to learn section topic representations, which we describe in Section~\ref{sec:topic-modelling}. The methods differ in how this learnt topic information is integrated.
The first, described in Section~\ref{sec:sideconstraints}, uses side constraints \cite{sennrich2016} to incorporate information about the section topic associated with each sentence. For the second method we use \newcite{kuang2018}'s cache-based neural model to provide the model with topic information and previous context within the boundaries of the current section. While our first method is simple to implement and train, the second provides more fine-grained information about each section within an article. %This section describes in more detail how we propose to obtain topic representations and the two methods we employ.
    
\subsection{Modelling Section Topics}\label{sec:topic-modelling}
    
While Wikipedia section headings are a useful indication of the section boundaries, they are not necessarily optimal for determining the granularity of different topics. As can be seen in Figure~\ref{fig:biography-example}, some headings are too general, (e.g.~``Career'' can refer to any occupation and the contents of this section can vary immensely between a sports person and a politician), whereas other headings are too specific to particular people and do not allow for any useful generalisation (e.g.~``Eclectic filmmaking and John Wick: 2009-present''). We therefore choose not to use the headings themselves to determine our topics, and instead use topic modelling to induce topics from article sections.
    
\paragraph{Training topic models}
As in \cite{lw2014}, we use Latent Dirichlet Allocation (LDA) \cite{blei2003} to train topic models on our monolingual biography data, taking into account section boundaries. The resulting topic models are then used to obtain a distribution over topics for each section in the parallel data, and the topic of the section is selected based on the topic with the highest probability. Each topic is associated with its own distribution over the language's vocabulary.

% Copied from related work section
LDA places certain assumptions on the generation of topic probabilities for each document, as well as on the generation of words given a topic.
These assumptions can be regulated with the two hyperparameters of LDA, alpha and beta, which control the distribution of topics in documents and of words in documents. Setting them to very small values encourages sparsity in the model (i.e.~it encourages the model to assign higher probabilities to only a few topics for each document and only some words in each topic). This is useful for MT as the topic models need to be able to differentiate between sections of different topics and provide the most salient words for the particular topic. Another important hyperparameter of LDA is the number of topics to be produced by the model. %This number can have a very significant effect on the quality of the topics. 
It is chosen empirically and depends on the sections to be modelled.

\paragraph{Aligning language-specific topics}
Since topic models for the two languages in the parallel data are trained separately, there is no direct correspondence between topics on the two sides. In practice, when translating new documents, only the text and topic on the source side is available and yet in our proposed cached model (cf.~Section~\ref{sec:cache-based-model}), words from the target side topics are used. It is therefore necessary to be able to predict a target topic using the topic of the source language.
Topics in the two languages are aligned by taking the co-occurrence counts of each French/Bulgarian/Chinese topic with each English topic in the parallel corpus. The topics that co-occur the most frequently are considered to be aligned. Finally, the English topic models are also used to obtain the most probable words for each topic. 
    
\subsection{Incorporating Topic Information}

\subsubsection{Side Constraints}\label{sec:sideconstraints}
A simple yet effective way of integrating information at the sentence level is to use side constraints \cite{sennrich2016}, which consists in prepending (or appending) an extra token to the input sentence associated with the feature to which you wish to bias the translation. These feature tokens are added to the vocabulary and treated as normal tokens. It has previously been used for a range of phenomena including politeness (in the original article) and domain adaptation \cite{kobus-etal-2017-domain,caswell-etal-2019-tagged}. 

We therefore prepend a special token to the beginning of the source sequence to represent the topic of the current source section (cf.~Figure~\ref{fig:side-constraints}). Our intuition is that the model will learn to associate the presence of each topic with related target vocabulary and in a way that provides more adapted information than topics trained at the document level.

\begin{figure}[!ht]
    \centering\small
    \scalebox{0.85}{
    \begin{tabular}{l}
    \toprule
    $<$topic64$>$ Elle a débuté en août 2013 avec ``My Student Teacher''. \\
    $<$topic91$>$ Le 21 mars 2014, NC.A sort son single digital ``Hello Baby''.\\
    $<$topic91$>$ Elle sort ensuite son premier mini-album le 9 avril. \\
    \bottomrule
    \end{tabular}}
    \caption{An example of topics used as side constraints.}
    \label{fig:side-constraints}
\end{figure}

\subsubsection{Cache-based Neural Model}\label{sec:cache-based-model}

Our second method draws inspiration from \newcite{lw2014} in their use of caches to represent section information. We adapt the model proposed by \newcite{kuang2018} to the Transformer \cite{vaswani2017}, introducing some small changes. \newcite{kuang2018}'s model uses two caches containing words relevant to the topic of the document and the preceding context respectively (termed ``topic cache'' and ``dynamic cache''). In our version of the model, the topic refers to the learned topic of the section rather than the document.

% Example of cache models from Radina's thesis
\begin{figure}[!ht]
\centering
\includegraphics[width=\linewidth]{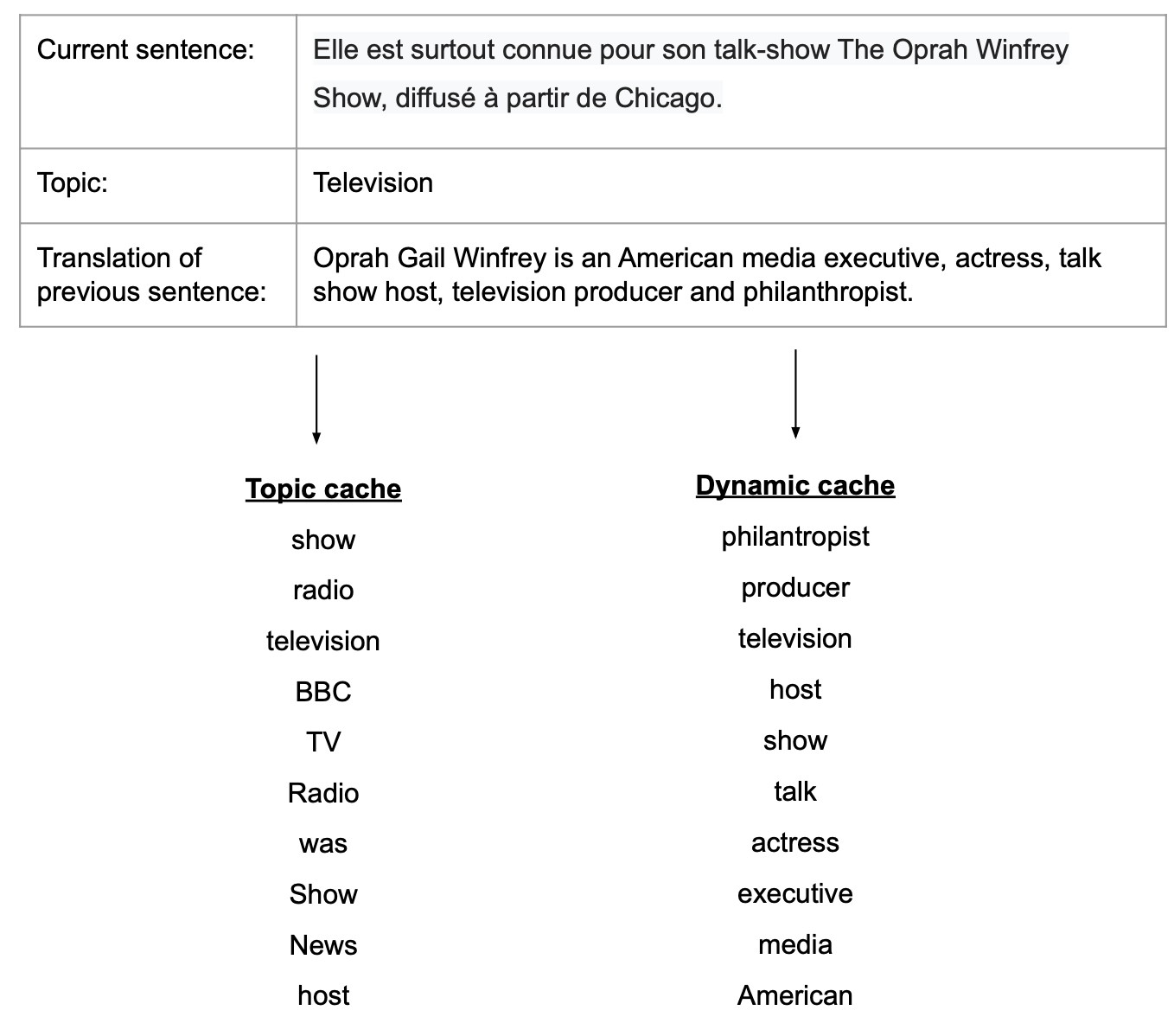}
\caption{An example of how the topic cache and dynamic contents are determined.}
\label{fig:cache-contents}
\end{figure}

As illustrated in Figure~\ref{fig:cache-contents}, the topic cache is loaded with words from the current topic, and the dynamic cache conveys information about the recent context preceding the current sentence. For a given sentence, the topic cache is loaded with the most probable words from the topic's distribution as learnt using the topic model, and the dynamic cache contains the set of unique content words\footnote{A list of grammatical (stopwords) was used to obtain only content words. We choose however to include pronouns and certain auxiliaries (e.g.~was, were) that could be useful indicators of tense choices (e.g.~for past events).} from previously translated sentences from the same section. 
The dynamic cache is updated as translation progresses to include the content words from the most recent preceding sentences. The cache is of a fixed size and if the length of the cache is exceeded, older context is removed to allow for newer context.% to be included. 
%Words are not duplicated within either cache. % Handled inline above
%If a context word is in the topic cache, it is not added again. If that word is already in the dynamic cache, the older entry is removed and the new entry is kept.

The caches are concatenated and passed to a neural cache model, which computes a probability distribution over the words. At timestep $t$, scores over cache words $y_c$ are computed based on the current decoder state $h_t$, the previous output from the NMT model $y_{<t}$ and the encoder context $c_e$ using a feedforward network $f_{cache}$:
\begin{align}
score\left(y_c|h_t,c_e,y_{<t}\right) &= f_{cache}\left(h_t, c_e, y_{<t}, y_c\right)
\end{align} 

The distribution over cache words is obtained by applying the softmax function:
\begin{align} 
p_{cache} &= softmax\left(score\left(y_c|h_t,c_e,y_{<t}\right)\right)
\end{align}

At each decoding step, this distribution is combined with $p_{\text{NMT}}$, the NMT model's distribution, using linear interpolation to obtain a final distribution over target words $y_t$:
\begin{align}
\small
g &= \sigma\left(f_{gate}\left(h_t, c_e, y_{<t}\right)\right) \\
p\left(y_t|y_{<t},c\right) &= g \cdot p_{\text{NMT}} + (1-g) \cdot p_{cache}
\end{align}

\begin{figure*}[!ht]
\centering
\includegraphics[scale=0.44]{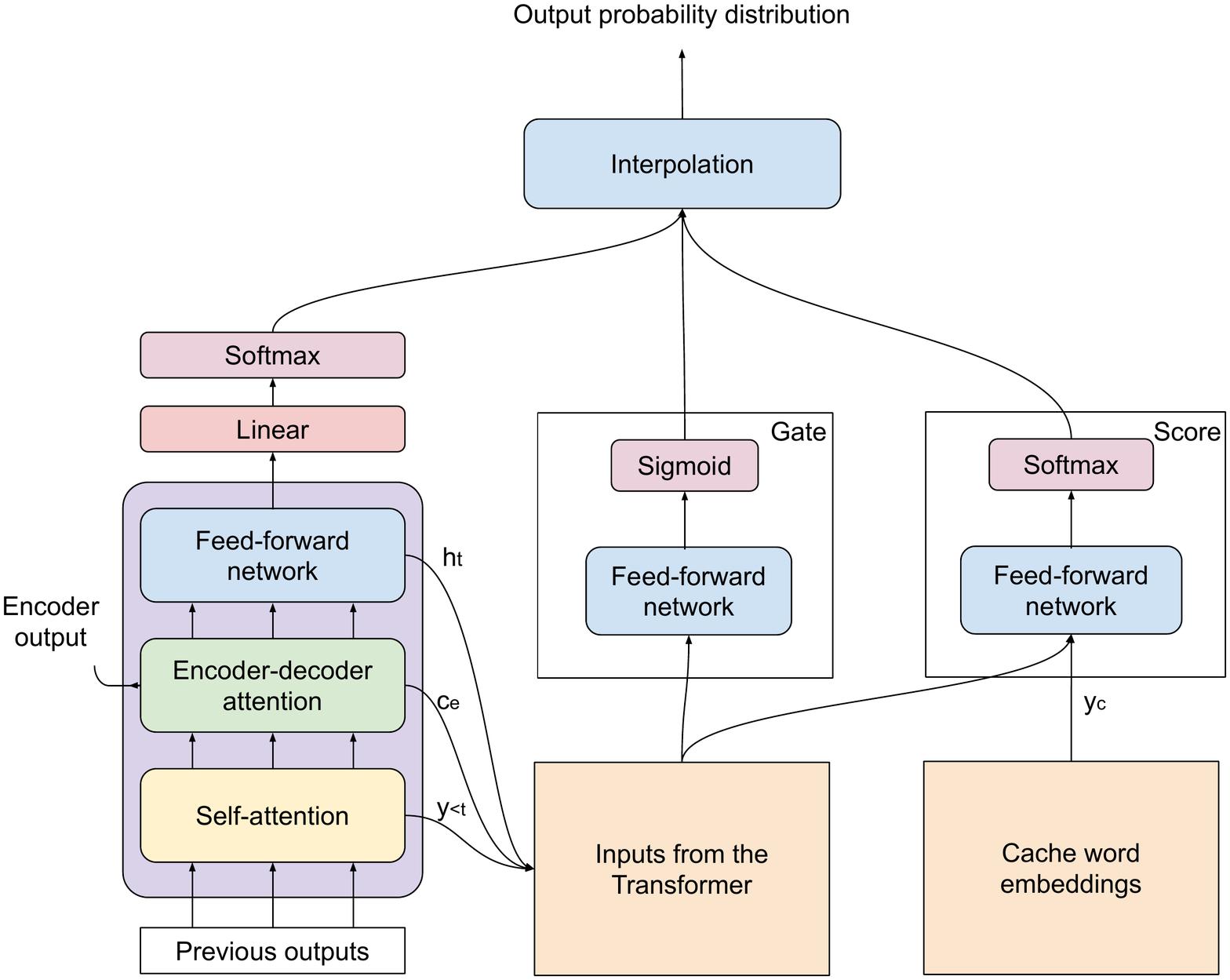}
\caption{The neural cache model integrated with the Transformer decoder. }
\label{fig:cache_model}
\end{figure*}

To adapt this method to the Transformer, we use the output of self-attention as a representation of the previously output words, the encoder-decoder attention to represent the source side context and the final feed-forward network as the current decoder state (see Figure \ref{fig:cache_model}).

% Cache-transformer schema

One difference between the present implementation and the cache model of \newcite{kuang2018} should be noted. While they update the dynamic cache at every timestep, providing the previously translated words from the same sentence, here the dynamic cache only contains words from preceding sentences. This is because the Transformer model has better access to output at preceding timesteps through self-attention. Furthermore, as the cache size is fixed, adding words from the current translation would reduce the length of preceding context that is available.
    
\section{Experiments}\label{sec:experiments}

We apply the two methods described in Section~\ref{sec:integrating-structure} to the translation of three languages (Fr, Zh and Bg) into English. We train and test the models using the Wikipedia biography data previously described in Section~\secref{sec:datasets}.

\subsection{Topic Models}
We use the monolingual datasets to train LDA topic models for the four languages (Fr, Zh, Bg, En). For each language we train two topic models: one that learns topics of sections within documents, and one that learns topics of the whole documents (for comparison purposes). The hyperparameters alpha and beta of LDA are set to 0.001 and 0.01 to encourage sparsity, following \cite{lw2014}. We set the number of topics to 100 for all models.

\subsection{MT Model Training}
    
\paragraph{Pre-training and Fine-tuning}

We train an out-of-domain Transformer-base model \cite{vaswani2017} for each language pair and use it as a basis to fine-tune all our models using our in-domain biography data. This pre-training ensures that all models are strong, having been trained on large quantities of data in addition to in-domain data. We also train an in-domain baseline by fine-tuning the pretrained model on our in-domain data, without additional information about document structure.% in order to ensure a fair comparison.

%%%% results table

\begin{table*}[!ht]
\centering\small
\begin{tabular}{llllllllll}
\toprule
& \multicolumn{3}{c}{Zh-En} & \multicolumn{3}{c}{Fr-En} & \multicolumn{3}{c}{Bg-En} \\ 
& En orig. & Zh orig. & all & En orig. & Fr orig. & all & En orig. & Bg orig. & all \\
\midrule
Out-of-domain baseline & 14.3 &  \textbf{11.6} &    13.1 & 45.3 &  46.1   &   45.7    & 22.0 &  21.9 &   22.0   \\
In-domain baseline  & 16.9 & 10.8 & 14.0  & \textbf{52.4} & \textbf{51.8}  &  \textbf{52.1}  & 24.4 & 22.0  &  24.2    \\ \midrule
Document-level side constraints   & 17.1 &  10.8 &   14.1   & 52.0* & \textbf{51.8} &   51.9  & \textbf{24.7}* & 23.1  &  24.5**  \\
Section-level side constraints & \textbf{17.3}** & 11.2* & \textbf{14.4}***   & 52.0 & 50.6*** &   51.3***  & \textbf{24.7}** &  \textbf{23.3}  &   \textbf{24.6}***      \\
Document-level cache-based   & 16.6 &  10.9  &  13.9  & 52.0 & 51.6   &  51.8   & 24.2  & 22.6 &   24.0   \\
Section-level cache-based  & 16.8 & 10.7 & 13.9  & 52.0  & 51.0** &   51.5**  & 24.3 & 22.5  &  24.1 \\
\bottomrule                     
\end{tabular}
\caption{\bleu score results. We distinguish between the two translation directions, depending on which language was the language of the original text (``orig.''), and also calculate the \bleu score on the concatenation of these two test sets (``all''). The highest score for each set is marked in bold and improvements that are statistically significant from those of the in-domain baseline are indicated with asterisks (* for $p\leq0.05$, ** for $p\leq0.01$ and ***$p\leq0.001$).}
\label{tab:results}
\end{table*}

\paragraph{Data and Preprocessing}

For the out-of-domain pretrained model, we use data from WMT \cite{barrault-etal-2019-findings} for Zh$\rightarrow$En ($\sim$24.2m parallel sentences) and Fr$\rightarrow$En ($\sim$39m parallel sentences), and data from Opus \citelanguageresource{opus} for Bg$\rightarrow$En ($\sim$44m parallel sentences). 
% 44021115 Bg-En
% 24252426 Zh-En
% 39048939 Fr-En
%
For all in-domain data we use the data described in Section~\secref{sec:datasets}. The data is tokenised and BPE segmented \cite{sennrich2015}. For cache models, words in the trained topics are also segmented using BPE. For the side constraints models we prepend a token representing the topic of each section.

\paragraph{Architecture and Settings}
The parameters of the Transformer are set to standard values \cite{vaswani2017}: the encoder and decoder have 6 stacked layers, the embedding size is 512 and the feed-forward network hidden layer dimensionality is 2048.
We use the Nematus toolkit \cite{nematus} for all models.

\paragraph{Cache model parameters}
For cache-based models, we fix the size of the two caches to 100 words each. The scoring feed-forward network has hidden dimensions 1000 and 500 and the gate feed-forward network has hidden dimensions 500 and 200, following the configurations reported in \cite{kuang2018}. The cache word embeddings are shared with the Transformer decoder. During training we provide the real topic of the target sentence for half of the training data, and for the other half we provide the topic projected from the source, in order for the model to learn from but not be over-reliant on gold (source) topics. During training we also use the real target side sentences to load the dynamic cache. At inference time, the topic is a projection from source to target and the dynamic cache is loaded with words from previously translated sentences.

\paragraph{Comparative Systems}
In order to assess the usefulness of document structure, we provide contrastive experiments whereby both methods are applied to the full Wikipedia articles, disregarding section boundaries, the difference being that topic models are trained on whole documents rather than individual document sections. We will refer to the models using section boundaries as ``section-level'' and to the ones that do not as ``document-level''.

\subsection{Results}\label{sec:results}

% results table moved to main.tex (just before "data and preprocessing")

We report \bleu scores \cite{papineni2002}, calculated using Sacre\bleu \cite{post-2018-call} on our tests sets for each of our models in Table~\ref{tab:results}. We also provide scores for the pretrained model (out-of-domain baseline) and the in-domain fine-tuned baseline. Improved results that are statistically significant\footnote{Calculated using bootstrap resampling using the Moses script \texttt{bootstrap-hypothesis-difference-significance.pl} on outputs and references tokenised using the WMT standard `13a' tokenisation as used in Sacre\bleu.} from the in-domain baseline results are indicated (* for $p\leq0.05$, ** for $p\leq0.01$ and ***$p\leq0.001$).  %We compare two versions of each structure-aware model, one using document-level topics and a second using section-level topics.

% Compare different scores between languages

% Baseline results and mention of the scores.
\paragraph{Effect of fine-tuning}
Fine-tuning the out-of-domain baseline gives improved results for all language pairs, with the greatest gain being seen for Fr-En with a +6.4 increase in \bleu. The smallest difference is seen for Zh-En, where fine-tuning actually degrades performance on the side of the test set that was originally in Chinese. Even for Fr and Bg, smaller gains are seen on this side of the test set.
\paragraph{Translation direction of test sets}
In fact, all language pairs and models show different results depending on the original language of the test sets - the scores are consistently higher for the En-originating test sets, than for the ones originating in Fr, Bg or Zh. This effect is particularly striking for Zh-En, where the difference reaches over 6 \bleu points for some models. One explanation for this is a difference in writing style. Text translated from another language has different properties from text originally written in the language. Since the majority of in-domain training data is text that was originally written in English, it is nor surprising that the models perform worse on the test sets originating in Fr/Bg/Zh.

\paragraph{Language-specific differences}
\bleu scores are very different depending on the language pair. Zh-En scores appear very low (from 10.8 to 17.3 \bleu), whereas Fr-En is characterised by very high scores (from 45.3 to 52.4 BLEU).
Upon closer inspection, it appears that one reason for the low scores achieved by the Zh-En models is that the references for the test sets are not always literal translations of the source, but contain many paraphrases and inexact translations (see Table~\ref{fig:ref-examples} for examples). %It is a limitation of \bleu that paraphrases do not score highly with this metric. 
The very high \bleu scores for Fr-En can be explained partly by the presence of fewer deviations in formulation, possibly due to the structural similarity of the two languages. The ease of this set is also aided by the presence of a number of short sentences making up lists of works (e.g.~books, musical compositions, etc.), which are relatively simple to translate.

\paragraph{Side-constraints vs.~cache model}
The comparative results of the two methods differ by language pair. For Bg-En, the side-constraints method shows a decisive advantage over the cache model method, with significantly higher \bleu scores (p$\leq$0.05).\footnote{With the exception of the Bg-orig. test set for which there is no significant difference between the two methods at the document level, which could be due to the fact that this test set is very small.} This is also observed for Zh-En, although to a lesser degree, and is only significant at the section level (p$\leq$0.05). The Fr-En results do not show any consistent pattern in the differences between the side-constraints and cache models. Overall it appears that the cache-based model does not offer a systematic advantage over the simpler side-constraints model, despite finer grained information being provided to the model.

\paragraph{Document- and section-level models}
According to \bleu, there is no systematic pattern between document- and section-level across all language pairs, and the scores depend on the method used. Whereas little difference is seen between the document- and section-level experiments for Zh-En and Bg-En, section-level information actually appears to degrade performance when compared to document-level information for Fr-En, and this for both the side constraints and cache-based methods. This effect is statistically significant for the Fr-original test set (p$\leq$0.05).

    % are the references not complete then? e.g. skip parts of the source?
    % omit subjects ...
    % ok so it's just because it's a difficult language pair to translate
    % yes
    % aslo the mismatch of domain 
    % the pretrain domian is new-comentaries ... 
    % but the scores are also low for the in-domain models?
    % we only give a tiny amount of in-domain parallel corpus compared with the pretrain-stage (million sentences)
    
    % wait, is the Zh->En data all back-translated?
    % nope they are real parallel sentences, ok forget about the back-translation part
    
    % ok.. i think it's weird that the out-of-domain system works best on the Zh-original test set though. means that there is a really big difference between texts originally in English and those originally in Chinese?
    % in terms of the writing style , yes ...
    % eg. in Zh-EN, people use idioms originally, which is hard to translate for both machine and human

\begin{figure*}[!ht]
    \centering\small
    \scalebox{0.95}{
    \begin{tabular}{lllllllll}
    \toprule
     \multicolumn{3}{c}{English} & \multicolumn{1}{l}{French} & \multicolumn{1}{l}{Bulgarian} & \multicolumn{1}{l}{Chinese} \\
     \#0  & \#32 &  \#63 & \#11 & \#3 & \#51\\
     \midrule
     band & film & married & il `he/it' & \bg{е} `is' & \zh{年} `year' \\
     their & films & was & Guerre `war' & \bg{ѝ} `her' & \zh{大学} `the University' \\
     was & director & died & Il `He/It' & \bg{тя} `she' & \zh{教授} `professor' \\
     album & directed & had & mondiale `world' & \bg{деца} `children' & \zh{毕业} `graduation' \\
     released & was & his & guerre `War' & \bg{има} `has' & \zh{学院} `College' \\
     they & producer & children & 1940 `1940' & \bg{майка} `mother' & \zh{研究} `the study'\\
     tour & his & wife & 1944 `1944' & \bg{баща} `father' & \zh{中国} `China' \\
     were & Film & son & puis `then' & \bg{години} `years' & \zh{担任} `Serve' \\
     first & He & daughter & Seconde `Second' & \bg{Тя} `She' & \zh{研究所} `graduate School' \\
     They & produced & They & 1945 `1945'  & \bg{дъщеря} `daughter' & \zh{获} `Gain' \\
     \bottomrule
    \end{tabular}}
    \caption{The top ten words in a selection of section-level topics.}
    \label{tab:topic_words}
\end{figure*}

\section{Analysis}\label{sec:analysis}
To gain more insight into the differences between the proposed models, we briefly analyse the cache model topics (Section~\ref{sec:analysis-topics}), and in Section~\ref{sec:manual-eval} provide a manual evaluation and some qualitative analysis of translations.

\subsection{Analysis of Topic Models}\label{sec:analysis-topics}

Table~\ref{tab:topic_words} shows the top ten most relevant words for selected topics. These lists show clearly that the topics are coherent, containing words relevant to the same subject matter. For instance, Topic 0 from the English topic model relates to sections concerning members of musical bands, and more specifically discussing the band's musical career. This is also true of the non-English topic models: Chinese Topic 51 is about higher education and academic career.

\begin{figure*}[!ht]
    \centering\small
    \scalebox{0.9}{
    \begin{tabular}{p{4.6cm}p{4.4cm}p{4.1cm}p{4cm}}
    \toprule
     Source  & 
     Reference & Document-aware model & Section-aware model\\
     \midrule
     \textbf{Soliste} internationale, Marielle Nordmann\ldots% partage son temps entre les concerts, l’enseignement et la création artistique. 
     & An international \textbf{soloist}, Marielle Nordmann\ldots% divides her time between concerts, teaching and artistic creation. 
     & An international classical \textbf{violist}, Marielle Nordmann\ldots% divides her time between concerts, teaching and artistic creation. 
     & An international \textbf{solo artist}, Marielle Nordmann\ldots %divides her time between concerts, teaching and artistic creation. 
     \\
    \midrule
    \zh{1919年，\textbf{参加了五四运动，加入少年中国学会。}} & 	In 1919, \textbf{he participated in the May Fourth Movement, China Youth Association.} & In 1919, \textbf{he took part in the May 4th Movement and joined the Young China Institute}. & \textbf{He joined the Boys' China Society} in 1919. \\
    \midrule
    \bg{\vspace{-0.32cm}Два дни по-късно \textbf{получава първото генералско звание - генерал-майор} и е назначен за командващ на 2-ра армия.} & On 14 September 1944 he \textbf{was promoted to Major General and} was given the command of the Bulgarian Second Army. & He was appointed commander of the 2nd Army two days later. & Two days later, he \textbf{received the first rank of major-general and} was appointed commander of the 2nd Army. \\
    % \midrule
    % %\bg{Албърт Питър Лоу () 
    % \ldots \bg{е канадски геолог, изследовател на Канада, спортист.} & %Albert Peter Low (May 24, 1861 - October 9, 1942)
    % \ldots \textbf{was} a Canadian geologist, explorer \textbf{and athlete}.
    % & %Albert Peter Lowe (born October 18, 1925) 
    % \ldots \textbf{was} a Canadian geologist, explorer, \textbf{and sportsman}. & %Albert Peter Law (born July 17, 1974) 
    % \ldots \textbf{is} a Canadian geologist, explorer of Canada. \\
    % \midrule
    % \bg{По-късно започва да се занимава със строителство и \textbf{дърворезба}.} & 	Later he began to deal with the particular construction with \textbf{wood carving}, shaping the profession quickly.	& He later became involved in construction and \textbf{forestry}. & He later became involved in construction and \textbf{woodwork}. \\
    \bottomrule
    \end{tabular}}
    \caption{Illustration of the types of differences between document-aware and section-aware side constraint model  outputs.}
    \label{tab:sentence_examples}
\end{figure*}

\subsection{Qualitative Analysis of Translations}\label{sec:manual-eval}
While BLEU scores provide a general impression of the performance of MT models, they do not give any insight into the specific strengths and weaknesses of the models. Therefore, we manually evaluate a sample of the outputs. We compare the two best performing models (document-level and section-level side constraints), randomly sampling sections from each of the test sets, and keeping the first 100 sentences from each as sets for manual evaluation. 

\begin{table}[!ht]
    \centering\small
    \scalebox{0.95}{
    \begin{tabular}{llrrrr}
    \toprule
     Lang. pair & Orig. & Better & Worse & Equal & Identical \\
     \midrule
    \multirow{2}{*}{Fr-En} & En & 15 & 14 & 22 & 49 \\
    & Fr & 13 & 16 & 33 & 38 \\
    \hline
    \multirow{2}{*}{Bg-En} & En & 22 & 17 & 23 & 38 \\
    & Bg & 24 & 24 & 32 & 20 \\
    \hline
    \multirow{2}{*}{Zh-En} & En & 22 & 9 & 34 & 35 \\
    & Zh & 9 & 11 & 65 & 15 \\
    \bottomrule
    \end{tabular}}
    \caption{Manual evaluation of 100 sentences per test set. Comparisons are classified in terms of the number of times the section-level translation is better, worse, equal or identical to the document-level output.}
    \label{tab:manual_analysis}
\end{table}

The results in Table~\ref{tab:manual_analysis} show that across all test sets, many sentences are translated identically by the two models. There is a further substantial number of sentences for which the two models achieve similar quality. Among the sentences that show a difference in quality between the two models, preference for either the section- or the document-level model depends on the original language of the set. Section-level models do better across the En-originating test sets, albeit to differing degrees: the effect is quite strong for Zh-En (22 better vs.~9 worse), but less strong for Bg-En (22 vs.~17) and negligible for Fr-En (15 vs.~14). For sets originating in Zh/Fr/Bg, section-level models perform similarly or worse than document-level models. This appears to show that section information is useful compared to document-level information, but only when translating from translationese (a pattern also seen in the \bleu scores). This effect also diminishes as the MT quality increases (as is the case with En-Fr, where the \bleu scores are particularly high). These results differ from those in \cite{lw2014}, where they do see improvements. However they do not test on the two translation directions (as the translationese debate in MT has only emerged in recent years). Moreover, it is possible that topic information is less useful for NMT models in high resource settings (as is the case for all three directions tested), as the quality of the models is already very good.
%\todo[inline]{provide a hypothesis about why this is. What is lacking is a final conclusion about the results - what do we learn from this? This could be of the form - we do not see the same conclusions as Louis and Webber, or we see similar things. Could it be that NMT is just too good for this to make a difference? Perhaps it would work better for low resource languages?}

Figure~\ref{tab:sentence_examples} shows examples of sentences which are translated differently by the two side-constraints models (more examples can be found in Appendix~\ref{app:outputs}). 
The main differences we observe for both models are in lexical choice, under-translations of some sentences, reformulations and differences in punctuation.
There does not appear to be a clear pattern in terms of the improvements or degradations seen by each model, and it is likely that the topic information is providing some domain adaptation effect, which is difficult to observe. A possible exception to this could be the first example in Table~\ref{tab:sentence_examples}, where \textit{solist} `soloist' is translated as \textit{violist} (i.e.~viola player), despite the woman described being a harp player. This sentence's document topic is music-related, and the model may have overfit to the topic.

\section{Conclusion and Future Work}\label{sec:conclusion}
In this paper we propose two methods to transfer into the framework of NMT \newcite{lw2014}'s idea of exploiting document structure when translating documents with regular and predictable structure. As in their work, we use topic modelling to model document sections, under the assumption that different sections within articles display different lexical properties, and compare the integration of this information using side constraints \cite{sennrich2016} to a more complex approach using cache mechanisms, adapted from \cite{kuang2018}.

We have created three parallel corpora of Wikipedia biographies (for En-Fr, En-Bg and En-Zh), structured into sections, as well as monolingual corpora for all four languages, all of which will be made freely available.

Our experiments using this data show that there are no consistent gains to be seen across all language directions for a particular model type, and using section-level information as opposed to document-level information does not systematically improve MT quality. These results, which are different from results found by \cite{lw2014} for SMT, suggest that while there are circumstances in which providing section (or document) topic information does help NMT through domain adaptation (shown by gains for certain subsets of the test set), in high-resource scenarios such as the ones tested here, this information is not systematically useful. A direction to look into in the future is how this type of information could help low-resource MT, as a way of injecting monolingual topic information in a more light-weight and efficient manner than most current techniques \cite{lample_XLM_2019,lample2017unsupervised}.

%
%We have presented two methods of incorporating structural information making use of the assumption that different sections within articles have different topics. 
%While our results are inconsistent and cannot serve as conclusive evidence for or against the usefulness of document structure in NMT, we are positive that further experimentation can shed more light on this.

\section{Acknowledgments}
We would like to thank Bonnie Webber, Annie Louis and Christian Hardmeier for their invaluable help and feedback on this project. This work was supported by funding from the European Union’s Horizon 2020 research  and  innovation  programme  under  grant agreement  No  825299  (GoURMET). It was also supported by the UK Engineering and Physical Sciences Research Council (EPSRC) fellowship grant EP/S001271/1 (MTStretch).

%\nocite{*}
\section{Bibliographical References}
\label{main:ref}

\bibliographystyle{lrec}
\bibliography{ref}

%\nocitelanguageresource{*}
\section{Language Resource References}
\label{lr:ref}
\bibliographystylelanguageresource{lrec}
\bibliographylanguageresource{lang_res}

\clearpage
\appendix
\onecolumn
\section{Examples from document-level and section-level side-constraints models}\label{app:outputs}

\begin{figure*}[!ht]
    \centering\small
    \scalebox{0.9}{
    \begin{tabular}{p{4.6cm}p{4.4cm}p{4.1cm}p{4cm}}
    \toprule
     Source  & 
     Reference & Document-aware model & Section-aware model\\
     \midrule
     \textbf{Soliste} internationale, Marielle Nordmann\ldots% partage son temps entre les concerts, l’enseignement et la création artistique. 
     & An international soloist, Marielle Nordmann\ldots% divides her time between concerts, teaching and artistic creation. 
     & An international classical \textbf{violist}, Marielle Nordmann\ldots% divides her time between concerts, teaching and artistic creation. 
     & An international \textbf{solo artist}, Marielle Nordmann\ldots %divides her time between concerts, teaching and artistic creation. 
     \\
     \midrule
     %Vladimir Shlyapnikoff, directeur de la Kapella de Saint-Petersbourg, 
     \ldots cité par l’abbé Abel Gaborit. \textbf{(dans ``Musica et Memoria'')}	& 
     %Vladimir Shlyapnikoff, director of the Saint Petersburg Court Chapel, 
     \ldots mentioned by abbot Abel Gaborit. \textbf{(in ``Musica et Memoria'')} & %Vladimir Shlyapnikoff, director of the Kapella of St. Petersburg, 
     \ldots quoted by Abel Gaborit. \textbf{(in ``Musica and Memoria'')}	& %Vladimir Shlyapnikoff, director of the Kapella de Saint-Petersbourg,
     \ldots quoted by Abel Gaborit.\\
     \midrule
     Entre 1960 et 1978, elle a \textbf{formé} le Trio Nordmann avec\ldots% le flûtiste André Guilbert et le violoncelliste Renaud Fontanarosa. 
     & Between 1960 and 1978, she \textbf{led} the Nordmann Trio with\ldots %flautist André Guilbert and cellist Renaud Fontanarosa. 
     & Between 1960 and 1978 she \textbf{formed} the Nordmann Trio\ldots% with the flutist André Guilbert and the cellist Renaud Fontanarosa. 
     & Between 1960 and 1978 she \textbf{trained} the Nordmann Trio %with flautist André Guilbert and cellist Renaud Fontanarosa. 
     \\
    \midrule
    \zh{相较于斯科特探险队的不幸，阿蒙森的探险\textbf{比较平顺}。} & In contrast to the misfortunes of Scott's team, Amundsen's trek proved \textbf{relatively smooth and uneventful}. & Compared with the Scott expedition's misfortune, Amundsen's expedition was \textbf{smoothness}. & Compared with the Scott expedition's misfortune, Ammonson's expedition was \textbf{relatively smooth.} \\
    \midrule
    \zh{要求\textbf{抛开关于自己处境的幻想}，也就是\textbf{要求抛开那需要幻想的处境}。} & To call \textbf{on them to give up their illusions about their condition} is to \textbf{call on them to give up a condition that requires illusions}. & To ask for an \textbf{abandoned fantasy about your situation}, that is, \textbf{a situation that requires fantasy}. & To ask for an \textbf{abandonment of the illusion of one's own situation}, that is, \textbf{to ask for an abandonment of the situation that requires fantasy}. \\
    \midrule
    \zh{1919年，\textbf{参加了五四运动，加入少年中国学会。}} & 	In 1919, \textbf{he participated in the May Fourth Movement, China Youth Association.} & In 1919, \textbf{he took part in the May 4th Movement and joined the Young China Institute}. & \textbf{He joined the Boys' China Society} in 1919. \\
    \midrule
    \bg{Два дни по-късно \textbf{получава първото генералско звание - генерал-майор} и е назначен за командващ на 2-ра армия.} & On 14 September 1944 he \textbf{was promoted to Major General and} was given the command of the Bulgarian Second Army. & He was appointed commander of the 2nd Army two days later. & Two days later, he \textbf{received the first rank of major-general and} was appointed commander of the 2nd Army. \\
    \midrule
    %\bg{Албърт Питър Лоу () 
    \ldots \bg{\textbf{е} канадски геолог, изследовател на Канада, \textbf{спортист}.} & %Albert Peter Low (May 24, 1861 - October 9, 1942)
    \ldots \textbf{was} a Canadian geologist, explorer \textbf{and athlete}.
    & %Albert Peter Lowe (born October 18, 1925) 
    \ldots \textbf{was} a Canadian geologist, explorer, \textbf{and sportsman}. & %Albert Peter Law (born July 17, 1974) 
    \ldots \textbf{is} a Canadian geologist, explorer of Canada. \\
    \midrule
    \bg{По-късно започва да се занимава със строителство и \textbf{дърворезба}.} & 	Later he began to deal with the particular construction with \textbf{wood carving}, shaping the profession quickly.	& He later became involved in construction and \textbf{forestry}. & He later became involved in construction and \textbf{woodwork}. \\
    \bottomrule
    \end{tabular}}
    \caption{Illustration of the types of differences between document-aware and section-aware side constraint model  outputs.}
    \label{tab:full-sentence_examples}
\end{figure*}

\end{document}